# Augmentations: An Insight into their Effectiveness on Convolution Neural Networks

Sabeesh Ethiraj[1], Bharath Kumar Bolla[2]

[1]Liverpool John Moores University, London
sabeesh90@yahoo.co.uk

[2]Salesforce, Hyderabad, India
bolla111@gmail.com

**Abstract.** Augmentations are the key factor in determining the performance of any neural network as they provide a model with a critical edge in boosting its performance. Their ability to boost a model's robustness depends on two factors, viz-a-viz, the model architecture, and the type of augmentations. Augmentations are very specific to a dataset, and it is not imperative that all kinds of augmentation would necessarily produce a positive effect on a model's performance. Hence there is a need to identify augmentations that perform consistently well across a variety of datasets and also remain invariant to the type of architecture, convolutions, and the number of parameters used. This paper evaluates the effect of parameters using 3x3 and depth-wise separable convolutions on different augmentation techniques on MNIST, FMNIST, and CIFAR10 datasets. Statistical Evidence shows that techniques such as Cutouts and Random horizontal flip were consistent on both parametrically low and high architectures. Depth-wise separable convolutions outperformed 3x3 convolutions at higher parameters due to their ability to create deeper networks. Augmentations resulted in bridging the accuracy gap between the 3x3 and depth-wise separable convolutions, thus establishing their role in model generalization. At higher number augmentations did not produce a significant change in performance. The synergistic effect of multiple augmentations at higher parameters, with antagonistic effect at lower parameters, was also evaluated. The work proves that a delicate balance between architectural supremacy and augmentations needs to be achieved to enhance a model's performance in any given deep learning task.

**Keywords:** Deep Learning, Depth-wise Separable convolutions, Global Average Pooling, Cutouts, Mixup, Augmentations, Augmentation paradox.

## 1 Introduction

Data augmentations have become a crucial step in the model building of any deep learning algorithm due to their ability to give a distinctive edge to a model's performance and robustness [1]. The efficiency of these augmentations largely depends on the type and the number of augmentations used in each scenario. The effectiveness of augmentations in improving a model's performance depends on model capacity, the number of training parameters, type of convolutions used, the model architecture, and the dataset used. Augmentations are very specific to the factors mentioned herewith, and it is not imperative that all kinds of augmentation would necessarily produce a positive effect on a model's performance. Furthermore, very few studies have evaluated the relationship between augmentations, model capacity, and types of convolutions used.

Data augmentations can range from simple techniques such as Rotation or random flip to complex techniques such as Mixup and Cutouts. The efficiency of these techniques in improving a model's performance and robustness is critical in the case of smaller architectures as they are ideal for deployment on Edge / Mobile devices due to the lesser model size and decreased training parameters. Not every deep learning task requires an architecture optimized on generic datasets like Imagenet. Hence, there is a need to build custom-made lightweight models that perform as efficiently as an over-parameterized architecture. One such technique to help achieve reduced model size is the utilization of Depth-wise Separable convolutions[2]. In this paper, experiments have been designed to answer the primary objectives mentioned below.

- To evaluate the relationship between model capacity and augmentation
- To evaluate the effect of model capacity on multiple augmentations
- To evaluate the effect of depth wise separable convolution on augmentation

## 2 Literature Review

Lately, work on augmentations and architectural efficiency has been the hallmark of research, making models more efficient and robust. The work done by showed that fine tuning layers of a pre trained network in the right proportion [3] resulted in better model performance than the baseline architecture itself suggesting that not all pre trained networks can be used in its nascent form. The research methods implemented in the paper have been implemented in various works summarized below.

### 2.1 Depth-wise Separable Convolutions, GAP & Architecture Fine-tuning

Depth-wise convolutions were first established in the Xception network [4] and were later incorporated in the MobileNet [2] architecture to build lighter models. Due to the mathematical efficiency, these convolutions help reduce the number of training parameters in contrast to a conventional 3x3 kernel. Traditional CNNs linearize the learned parameters by creating a single dimension vector at the end of all convolutional blocks. However, it was shown that this compromises a network's ability to localize the features extracted by the preceding convolutional blocks [5]. It was also shown that [6] that earlier layers capture only low-level features while the higher layers capture task-specific features, which need to be preserved to retain much of the information. This was made possible by the concept of Global Average Pooling, wherein the information learned by the convolutions is condensed into a single dimension vector. The work done by [3] has also stressed the importance of Depth-wise convolutions and Global average pooling layer wherein deep neural networks were trained using these techniques, and a significant improvement in model performance was observed despite the reduction in the number of parameters. The techniques have been widely used in industrial setting as in the case of detecting minor faults in case of gear box anomalies [7], where in retaining special dimension of features is very important, considering the minor nature of these faults and a high probability of missing them during regular convolutions. A similar study in the classification of teeth category was done where in the results of max pooling was compared with that of average pooling [8]. The efficiency of these architectural modifications was further validated as recently as in the work done by [9] [10] where custom architectures were built and were found to be superior to pre trained architectures trained on Imagenet. The effect of augmentations on these networks were also studied and in some cases were found to have a paradoxical effect on the output of the model. In addition to conventional techniques, state-of-the-art techniques such as Blurpool, Mixup [11] were shown to produce higher model accuracy than conventional methodologies.

### 2.2 Data Augmentation

Work on Regularization functions was done as early as 1995 to make models more robust, such as the radial and hyper basis functions [12]. These focus on a better approximation of the losses. Bayesian regularized ANNs [13] were more robust than conventional regularization techniques, which work on the mathematical principle of converting non-Linear regression into ridge regression, eliminating the need for lengthy cross-validation. State-of-the-art results were obtained on CIFAR-10 and CIFAR-100 datasets using Cutouts [14] to make models more robust and generalizable. Similarly, [15] the concept of mixup wherein combining the input and target variables resulting in a completely new virtual example resulted in higher model performance due to increased generalization. Even conventional baseline augmentations [1] have been shown to perform complex augmentation techniques such as GANs in augmenting the training samples. Combining self-supervision learning and transfer learning has also been shown to boost model performance [16] when no label is provided in some instances to enable powerful learning of the feature representations that are not biased. However, the effects of augmentations are not simply restricted to the improvement of model performance alone, as seen in the work done by [17], where other properties of augmentation such as test-time, resolution impact, and curriculum learning were studied. Though augmentations generally improve model performance most of the time, augmentation may not necessarily and always positively affect a neural network. They do have their shortcomings, as described in work done by [18].

## 3   Research Methodology

The methodology focuses on studying the effect of augmentations on different architectures (varying parameters) across three different datasets (CIFAR10, FMNIST and MNIST consisting of 50000 training and 10000 validation samples) using 3x3 and depth wise separable convolutions. Augmentation techniques have been chosen judiciously to perform a wide variety of transformations to the input images so that the effect of these techniques would be more explainable from the research perspective. The augmentations have been performed using Pytorch and the Mosaic ML library[19].

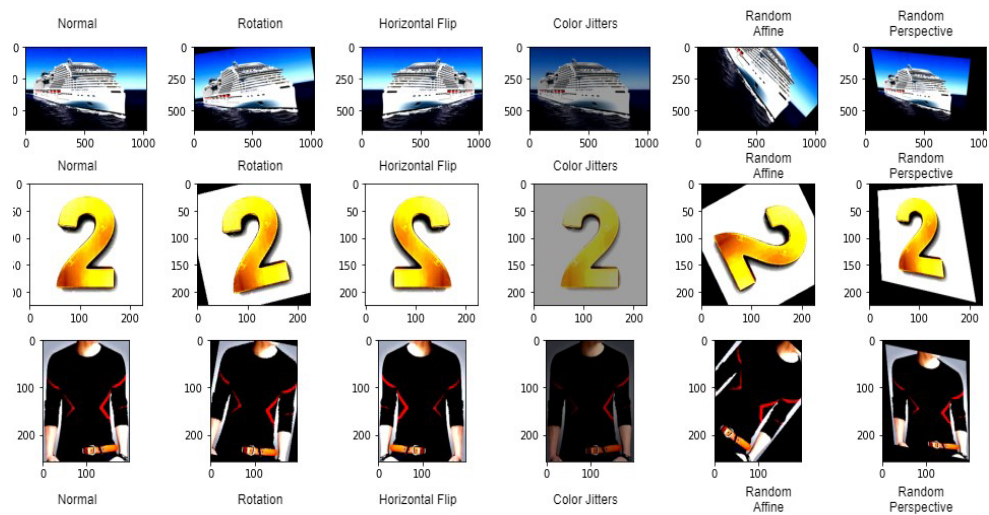

Figure 1. A representation of various augmentation techniques

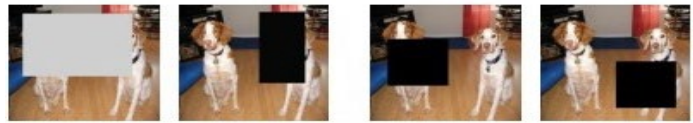

Figure 2. Cutout

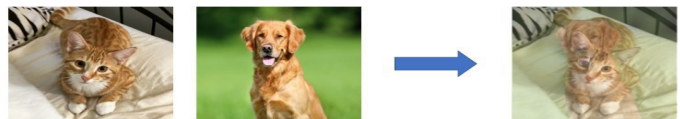

Figure 3. Mixup

### 3.1   Depth-wise Separable Convolutions

The mathematical intuition behind the depth wise convolutions is reducing the number of training parameters for the same number of features extracted by a 3x3 kernel. This is achieved by combining depth-wise channel separation and a 3x3 kernel followed by a point-wise convolution to summate the separated features learned in the previous step.

### 3.2 Global Average Pooling

Global average Pooling (GAP) is performed by taking the average of all the neurons/pixels in each output channel of the last convolutional layer resulting in a linear vector.

### 3.3 Random Rotation and Random Horizontal Flip

Random Rotation of 10 degrees (Figure 1) was chosen and applied uniformly on all the datasets to keep the variation constant across datasets. The rotation was kept minimal at 10 degrees to avoid any significant distortion of the original distribution. In case of random horizontal flip, the augmentation is applied with a probability of 0.5, where there is a lateral rotation of the images.

### 3.4 Random Affine and Random Perspective

Random Affine (Figure 1) is a combination of rotation and a random amount of translation along the width and the height of the image as defined by the model's hyperparameters. Random perspective performs a random perspective transformation of the input image along all the three axes.

### 3.5 Cutout

The Cutout is a regularisation or augmentation technique in which pixels from an input image are clipped. Random 8x8 masks (Figure 2 – Sample) have been clipped in the experiments to avoid any extreme influence of both smaller and larger Cutouts which can affect model performance significantly.

### 3.6 Mixup

Mixup is a regularization technique in different input samples and their target labels to create a different set of virtual training examples, as shown in Figure 3. A hyperparameter δ controls the mixup. The mathematical formula for mixup is shown in Equation 1 where $\hat{x}$ and $\hat{y}$ are new virtual distributions created from the original distribution

$$\hat{x} = \delta x_i + (1-\delta) x_j$$
$$\hat{y} = \delta y_i + (1-\delta) y_j$$

**Equation 1.** Virtual distribution

### 3.7 Loss Function

The loss function used here is the cross-entropy loss as this is a multiclass classification.

$$CE = -\sum_{i}^{C} t_i log(s_i)$$

**Equation 2.** Cross-Entropy Loss

### 3.8 Evaluation Metrics

Validation Accuracy and the percentage of accuracy change from the baseline model for every architecture are used as evaluation metrics.

### 3.9 Model Architecture

**Table 1.** Model Architectures

| Architectures incorporating Global Average Pooling across datasets | | |
|---|---|---|
| MNIST | Fashion MNIST | CIFAR-10 |
| 1.5K - 1560 | **1.5K DW**   - **1560**<br>5.7K NDW   - 5722<br>**5.6K DW**   - **5626**<br>7.8K NDW   - 7,777<br>**7.6K DW**   - **7,621**<br>25K NDW   - 25,154<br>**25K DW**   - **24,644**<br>140K NDW   - 142,930<br>**140K DW**   - **143,118**<br>600K NDW   - 600,575<br>**600K DW**   - **599,625** | 5.8K NDW   - 5,886<br>7.9K NDW   - 7,921<br>25K NDW   - 25,298<br>**25K DW**   - **24,788**<br>140K NDW   - 143,218<br>**140K DW**   - **143,406**<br>340K NDW   - 340,010<br>**340K DW**   - **344,508**<br>600K NDW   - 590,378<br>**600K DW**   - **599,913**<br>1M NDW   - 1,181,970<br>**1M DW**   - **1,159,474** |

(* NDW – non depth wise (3x3) / DW - Depth-wise Separable convolutions)

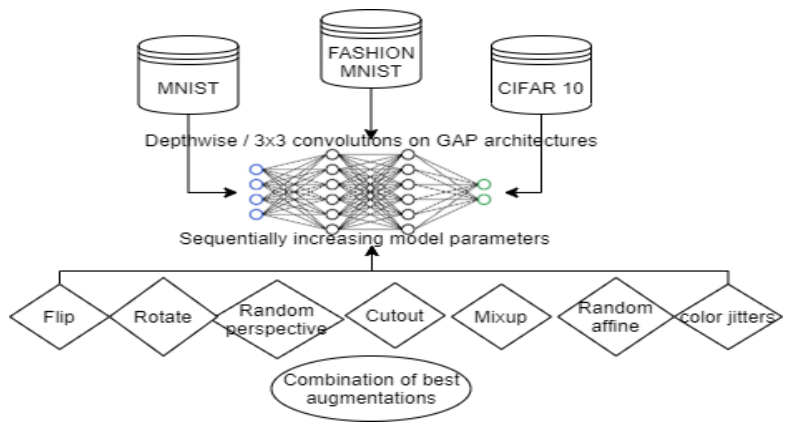

Figure 4. Proposed Methodology

Architectures (Table 1) have been built by sequentially reducing the number of parameters using the concepts of Depth Wise Convolutions and Global Average Pooling followed by application of various augmentations (Figure 4).

## 4 Results

The validation accuracies of different architectures (parameters) have been summarized in Tables 1,2,3, and 4 on both the FMNIST and CIFAR 10 datasets.

Table 1. Accuracies of 3x3 Convolutions – FMNIST

| Techniques | 5.7K | 7.8K | 25K | 140K | 600K |
|---|---|---|---|---|---|
| Baseline | 90.12 | 90.6 | 91.81 | 93.55 | 94.19 |
| Cutouts | 91 | 91.51 | 92.45 | 93.87 | 94.7 |
| Mixup | 90.03 | 90.29 | 91.65 | 94.17 | 94.7 |
| Random Rotation | 90.17 | 90.34 | 91.64 | 93.57 | 94.39 |
| Random horizontal flip | 91.07 | 91.23 | 92.17 | 93.81 | 94.72 |
| Color jitters | 88.04 | 88.69 | 88.61 | 91.38 | 92.17 |
| Random Affine | 86.38 | 86.52 | 89.12 | 92.36 | 93.51 |
| Random perspective | 88.72 | 89.17 | 90.89 | 92.83 | 94.06 |

Table 2.Accuracies of Depth-wise Separable Convolutions - FMNIST

| Techniques | 1.5K | 5.6K | 7.6K | 25K | 140K | 600K |
|---|---|---|---|---|---|---|
| Baseline | 87.39 | 89.32 | 90.03 | 91.28 | 93.36 | 94.19 |
| Cutouts | 88.74 | 91.09 | 91.22 | 92.18 | 93.82 | 94.85 |
| Mixup | 86.49 | 89.69 | 89.9 | 91.71 | 93.99 | 94.84 |
| Random Rotation | 87.51 | 90.24 | 90.47 | 91.29 | 93.45 | 94.42 |
| Random horizontal flip | 87.57 | 90.37 | 91.27 | 91.86 | 93.57 | 94.92 |
| Color jitters | 84.7 | 88 | 88.09 | 88.51 | 91.28 | 92.13 |
| Random Affine | 83.23 | 85.85 | 86.68 | 88.29 | 92.41 | 93.57 |
| Random perspective | 85.41 | 88.29 | 88.59 | 90.31 | 92.86 | 94.43 |

Table 3.Accuracies of 3x3 Convolutions - CIFAR10

| Techniques | 5.7K | 7.8K | 25k | 140k | 340K | 590K | 1M |
|---|---|---|---|---|---|---|---|
| Baseline | 70.12 | 72.15 | 73.67 | 82.79 | 85.49 | 86.3 | 85.69 |
| Cutouts | 70.89 | 72.06 | 75.91 | 84.53 | 87.04 | 87.6 | 87.49 |
| Mixup | 71.62 | 73.33 | 76.33 | 84.81 | 87.62 | 87.54 | 88.21 |
| Random Rotation | 70.35 | 72.11 | 74.97 | 83.02 | 85.84 | 86.45 | 86.16 |
| Random horizontal flip | 71.58 | 73.32 | 77.41 | 84.9 | 87.76 | 88.53 | 88.12 |
| Color jitters | 69.2 | 68.79 | 71.96 | 79.91 | 82.92 | 83.93 | 83.31 |
| Random Affine | 59.93 | 60.61 | 68.39 | 81.13 | 84.76 | 85.55 | 85.17 |
| Random perspective | 67.31 | 67.85 | 74.55 | 82.7 | 86.71 | 87.66 | 86.83 |

Table 4. Accuracies of Depth-wise Separable Convolutions - CIFAR10

| Techniques | 25k | 140K | 340K | 590K | 1M |
|---|---|---|---|---|---|
| Baseline | 70.75 | 81.06 | 83.7 | 86.7 | 85.96 |
| Cutouts | 72.49 | 82.56 | 85.36 | 87.85 | 87.89 |
| Mixup | 72.85 | 83.45 | 86.14 | 88.86 | 88.55 |
| Random Rotation | 72.1 | 81.34 | 83.81 | 87.62 | 86.71 |
| Random horizontal flip | 74.66 | 84.04 | 86.72 | 89.12 | 89.01 |
| Color jitters | 70.89 | 78.98 | 81.09 | 84.2 | 84.17 |
| Random Affine | 61.8 | 79.43 | 82.58 | 86.74 | 86.3 |
| Random perspective | 69.95 | 81.34 | 84.77 | 88.38 | 86.38 |

## 4.1 Augmentations on Depth-Wise Separable Convolutions

**Equivocal performances on Architectural Saturation.** On the FMNIST dataset, baseline architectures with fewer parameters and 3x3 convolutions performed better than depth wise separable convolutions due to a 3x3 convolution's improved feature extraction. However, on the incorporation of augmentation techniques, the difference in validation accuracies between the depth-wise Conv and 3x3 Conv architectures diminishes considerably ("*Diminishing differences*") with **equivocal performances** by both types of convolutions in most cases (Table 5 and Table 6) as the architecture approaches **saturation**, beyond which there is no significant improvement in accuracy even with augmentations.

Table 5. The difference in accuracies (3x3 Conv - depth wise conv) – FMNIST

| Techniques | 5.5K | 7.8K | 25K | 140K | 600K |
|---|---|---|---|---|---|
| Baseline | 0.8 | 0.57 | 0.53 | 0.19 | 0 |
| Cutouts | -0.09 | 0.29 | 0.27 | 0.05 | -0.15 |
| Mixup | 0.34 | 0.39 | -0.06 | 0.18 | -0.14 |
| Random Rotation | -0.07 | -0.13 | 0.35 | 0.12 | -0.03 |
| Random horizontal flip | 0.7 | -0.04 | 0.31 | 0.24 | -0.2 |
| Colour jitters | 0.04 | 0.6 | 0.1 | 0.1 | 0.04 |
| Random affine | 0.53 | -0.16 | 0.83 | -0.05 | -0.06 |
| Random perspective | 0.43 | 0.58 | 0.58 | -0.03 | -0.37 |

**Additive effect of Model capacity in Depth wise convolutions**. The above phenomenon of diminishing differences is not seen in the CIFAR-10 experiments suggesting that there is scope for improvement in model performance by fine-tuning the layers. However, with **a higher number of parameters (>600K), depth wise convolutions perform better than 3x3 convolutions as they enable a neural network to go deeper in terms of the number of convolutional layers.**

Table 6. Difference in accuracies (3x3 conv - depth wise Conv) - CIFAR10

| Techniques | 25K | 140K | 340K | 600K | 1M |
|---|---|---|---|---|---|
| Baseline | 2.92 | 1.73 | 1.79 | -0.4 | -0.27 |
| Cutouts | 3.42 | 1.97 | 1.68 | -0.25 | -0.4 |
| Mixup | 3.48 | 1.36 | 1.48 | -1.32 | -0.34 |
| Random Rotation | 2.87 | 1.68 | 2.03 | -1.17 | -0.55 |
| Random horizontal flip | 2.75 | 0.86 | 1.04 | -0.59 | -0.89 |
| Colour jitters | 1.07 | 0.93 | 1.83 | -0.27 | -0.86 |
| Random affine | 6.59 | 1.7 | 2.18 | -1.19 | -1.13 |
| Random perspective | 4.6 | 1.36 | 1.94 | -0.72 | 0.45 |

### 4.2 Consistency of Augmentations on Architectural Diversity

Augmentation techniques have been applied to different architectures and a relative ranking score was given to each of these techniques based on the average change in accuracy and standard deviation. A higher ranking was given to augmentations with the least standard deviation and higher gain in accuracy (Table 7). It was observed that Cutouts and a simple technique such as Random Horizontal Flip performed consistently superior to other techniques. It remained invariant to change in model capacity, architectural depth, and convolutions as evident from the least standard deviation and highest change in accuracy. On the CIFAR-10 dataset, mixup achieved the highest accuracy which is attributed to the wide distribution of classes. At the same time, the same technique and random horizontal flip decreased model performance on the MNIST dataset. Random affine, colour jitters, and random perspective negatively impacted (Augmentation paradox) on the accuracy on both the datasets.

Table 7. Change in accuracies / Change in Standard Deviation - Augmentation

| Aug | FMNIST | | | CIFAR 10 | | | MNIST |
|---|---|---|---|---|---|---|---|
| | STD | Change in Acc | Rank | STD | Change in Acc | Rank | Change in Acc |
| Cutouts | **0.47** | **0.96** | 1 | 0.76 | 1.78 | 2 | **0.1** |
| Mixup | 0.53 | 0.15 | 4 | **0.59** | **2.59** | 1 | -0.5 |
| Rand Rotate | 0.34 | 0.17 | 3 | 0.61 | 0.65 | 3 | 0.46 |
| Rand HFlip | 0.38 | 0.67 | 2 | **1.10** | **3.21** | 1 | -0.81 |
| Color jitters | 0.54 | -2.41 | 5 | 1.13 | -2.56 | 5 | 0.74 |
| Rand affine | 1.52 | -2.81 | 6 | 5.87 | -4.80 | 6 | -2 |
| Rand perspective | 0.69 | -1.04 | 7 | 2.35 | -0.14 | 4 | -0.99 |

### 4.3 The superiority of parameters over augmentations

The difference in accuracy gains/losses across the various augmentation strategies reduces as the number of model parameters increases. This tendency is particularly pronounced in augmentation strategies that degrade model performance. This is seen statistically (Figure 5), where higher architectures have lower standard deviations of accuracies. This phenomenon is indicative of the following hypothesis. Positive augmentations such as Cutouts, Mixup, Random rotation, and random horizontal flip have no meaningful influence on accuracy at larger parameter counts (low SD). In contrast, the performance of negative augmentations such as Color Jitters, Random affine, and Random perspective improves with the number of parameters. The neurons' enhanced learning capacity mitigates the effect of negative augmentation strategies at higher settings. In contrast, at lesser number of parameters, the effect of negative augmentation is exaggerated on both datasets as evident by the increased SD.

### 4.4 Combining Augmentations

Augmentations that positively affect the model performance in terms of accuracy were combined in varying combinations. At higher number of parameters combining different augmentations resulted in a synergistic effect. However, an antagonistic effect was observed at a lesser number of parameters. This can be attributed to models' relatively lesser learning capability with fewer parameters. The observations are summarized in Tables 8, 9, 10, and 11.

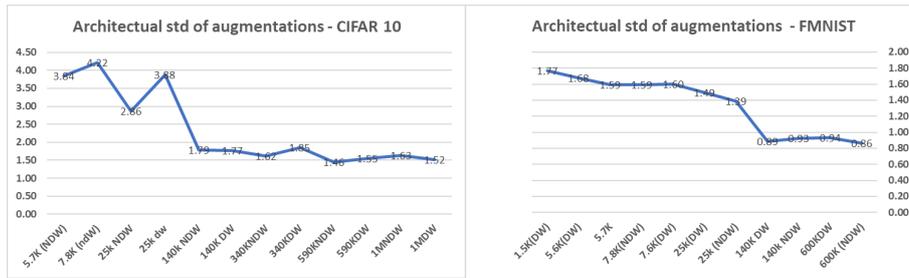

Figure 5. Standard deviation of changes in accuracy for various techniques

Table 8. Effect of Combining Augmentations - MNIST

| Models (1.5K params) - MNIST | Accuracy | Effect |
|---|---|---|
| 1. Baseline | 98.35 | Baseline |
| 2. Cutout +Random Rotation | 98.73 | Synergist to 1 |
| 3. Cutout + Random Rotation + Colour Jitters | 98.23 | Antagonist to 1 & 2 |
| 4. Cutout + Colour jitters | 98.9 | Synergist to 1,2 & 3 |

Table 9. Effect of Combining Augmentations - FMNIST

| Architecture | Accuracy w.r.t Augmentations - FMNIST | Effect |
|---|---|---|
| 140k 3x3 convolutions | 1.Baseline  – 93.55<br>2.Cutout + Horizontal flip  – 94.55<br>3. Cutout + Horizontal flip + mixup – 94.22 | Baseline<br>Synergist to 1<br>Antagonist to 2 |
| 140k depth wise convs | 1.Baseline  – 93.36<br>2.Cutout + Horizontal flip  – 94.48<br>3. Cutout + Horizontal flip + mixup – 93.71 | Baseline<br>Synergist to 1<br>Antagonist to 2 |
| 600 K | 1.Baseline  – 94.19 | Baseline |

| 3x3 convolutions | 2.Cutout + Horizontal flip           – 94.98<br>3. Cutout + Horizontal flip + mixup – 95.06 | Synergist to 1<br>Synergist to 1 and 2 |
|---|---|---|
| 6000K<br>depth wise convs | 1.Baseline                           – 94.19<br>2.Cutout + Horizontal flip           – 94.97<br>3. Cutout + Horizontal flip + mixup – 95.22 | Baseline<br>Synergist to 1<br>Synergist to 1 and 2 |

Table 10. Effect of Combining Augmentations - CIFAR10 – Lower architectures

| Sl | Techniques | 5.7K NDW | 7.8K NDW | 25k NDW | 25k DW | 140k NDW | 140K DW |
|---|---|---|---|---|---|---|---|
| 1 | Baseline | 70.12 | 72.15 | 73.67 | 70.75 | 82.79 | 81.06 |
| 2 | MU + RHP | 72.51 | 73.88 | 78.05 | 74.71 | 86.55 | 85.65 |
| 3 | MU + RHP + CO | 71.46 | 73.93 | 77.59 | 75.02 | 86.64 | 86.64 |
| 4 | Effect of 3 w.r.t 2 | **Ant** | **Syn** | **Ant** | **Syn** | **Syn** | **Ant** |

Table 11. Effect of Combining Augmentations - CIFAR10 - Higher architectures

| Sl No | Techniques | 340K NDW | 340 K DW | 590 K NDW | 590 K DW | 1M NDW | 1M DW |
|---|---|---|---|---|---|---|---|
| 1 | Baseline | 85.49 | 83.7 | 86.3 | 86.7 | 85.69 | 85.96 |
| 2 | MU + RHP | 89.72 | 88.51 | 90.05 | 91.88 | 89.76 | 90.58 |
| 3 | MU + RHP + CO | 89.92 | 88.75 | 90.69 | 91.98 | 90.2 | 90.98 |
| 4 | Effect of 3 w.r.t 2 | **Syn** | **Syn** | **Syn** | **Syn** | **Syn** | **Syn** |

## 5 Conclusion

The focus area of research in this paper has primarily been evaluating various augmentation techniques and arriving at an understanding of how model capacity and depth wise convolutions affect the outcome of an augmentation. The work has identified a new direction in appreciating those consistently invariant techniques and would apply them across a wide variety of datasets. Furthermore, this is the first study of its kind to unravel the relationship that exists between depth-wise convolutions, model capacity, and augmentations across a wide variety of standard datasets. The conclusions of the experiments are summarized below.

Augmentations such as Cutout, Random Horizontal flip, and Random Rotation performed consistently across all architectures. Considering the trade-off among training time, mathematical computational time, and model accuracy, it is suggested that a simple technique such as random horizontal flip, which performs equally well, may be used as a baseline augmentation. Further, these techniques were invariant to the number of parameters and the type of convolutions used, hence making them ideal for deployment on other real-life datasets. Combining augmentations worked well on over-parameterized architectures with the synergistic effect seen in all the cases. Depth wise separable Convolutions were effective on a higher number of parameters as they gave the ability of a model to go deeper and hence outperformed models with lesser parameters. Though on lesser parameterized architectures, 3x3 performed better, the application of augmentations bridged the accuracy gap between these architectures.